# What an RL agent relearns, retains and overlooks when its world is patched

Ding Wang[1,2], Xingyu Guo[1,2], Junru Jin[3], Jianbo Qiao[3], Shu Wu[1,2,*] and Qiang Liu[1,2,*]

[1]New Laboratory of Pattern Recognition (NLPR), State Key Laboratory of Multimodal Artificial Intelligence Systems (MAIS), Institute of Automation, Chinese Academy of Sciences, Beijing, China.

[2]School of Artificial Intelligence, University of Chinese Academy of Sciences, Beijing, China.

[3]School of Software, Shandong University, Jinan, China.

[*]Corresponding authors: Shu Wu and Qiang Liu (qiang.liu@nlpr.ia.ac.cn)

## Abstract

Reinforcement learning (RL) agents are trained in one version of a world and deployed in the next, yet how a policy absorbs a real change is rarely measured, because real changes cannot be undone. In a game under active development they can be: every update (patch) is a documented list of rule edits, each of which can be reverted. Here we present SpireZero, a small agent that learns *Slay the Spire 2* from its own play on the released game code, and follow it through real updates. SpireZero wins 73% of games at the highest difficulty, matching an independent, more heavily engineered agent. Small learning steps keep it improving over billions of decisions without loss of plasticity. One update, traced to a single card, was repaired within hours without loss of earlier skill; another, spread over 40 cards, only slowly. Both agents share one limit: abandoned options are rarely found again, and strategies that need several cards together are rarely found at all.

## Introduction

Reinforcement learning systems are rarely deployed into a world that stays still [1, 2]. Recommendation platforms, controllers and language agents all face rules and incentives that shift as software and people change around them. Continual-learning research has studied this problem mostly with synthetic changes, such as permuted pixels, drifting physics constants or sequences of unrelated tasks. Two failures dominate that literature: networks trained for a long time lose the ability to learn anything new (loss of plasticity) [3–5], and learning new rules erases old competence (catastrophic forgetting) [6]. What synthetic benchmarks cannot show is how a strong policy responds to the kind of change that real software undergoes, in which a developer edits a few dozen rules at once and nobody knows in advance which of them matters.

A game under active development offers that missing experiment [7, 8]. Every few weeks its developers ship an update (a patch) that changes a known set of cards, items and enemies. Because each game is fixed by its random seed, the same policy can play exactly the same games before and after an update, and any single change can be switched back to measure what it alone did (Fig. 1). *Slay the Spire 2*, a single-player card game in which a player builds a deck while fighting through three acts, is a demanding example. One game, called a run, consists of several hundred decisions across combat, deck building, route planning, shops, rest sites and events, and it ends with two consecutive boss fights in which choices made 40 floors earlier pay off or fail. Even the strongest frontier language models, tested on a frozen build of the game, win only about one run in three at the highest difficulty [9].

Here we introduce SpireZero, an agent trained with proximal policy optimization (PPO) [10] on the released code of the game, and we follow six copies of it, one for each playable character, through real updates. A small network that learns only from its own games reaches the level of a concurrent, much more heavily engineered system [11]. It keeps improving over billions of decisions because its learning steps are far smaller than is customary, and with such steps the network does not age. Two consecutive real updates can each be taken apart rule by rule, yet the agent absorbed one in hours and the other only slowly. From the outcome of its games alone, the agent learns to trade health, gold and cards across a whole run. Finally, a comparison with expert humans and with the independent agent shows where learning from one's own experience stops: an option that has been abandoned is rediscovered 10 times more slowly than it was dropped, and strategies that pay off only when several cards are combined are almost never found.

## Results

### A small agent in a game that keeps changing

SpireZero plays the real game. Our environment runs the released game code with only the graphics removed, so every decision a player makes is put to the agent as a choice among the legal options, and nothing is decided on its behalf (Fig. 1a, b). To confirm that this environment is the game, we replayed recorded runs inside the released game: what the agent saw and the options it was offered matched at every decision (Methods). The agent sees what a player with perfect memory would see: the cards, relics, potions, enemies and map on screen, but not the order of the draw pile or the contents of unopened rewards. It represents these as an unordered set of tokens, passes them through a four-layer Set Transformer, scores each legal action and estimates how the run will end (Fig. 1b). Apart from a lookup table that gives each name its own vector, the network has about five million weights. It uses no human games, no language model, no search at decision time and no hand-written knowledge of the game.

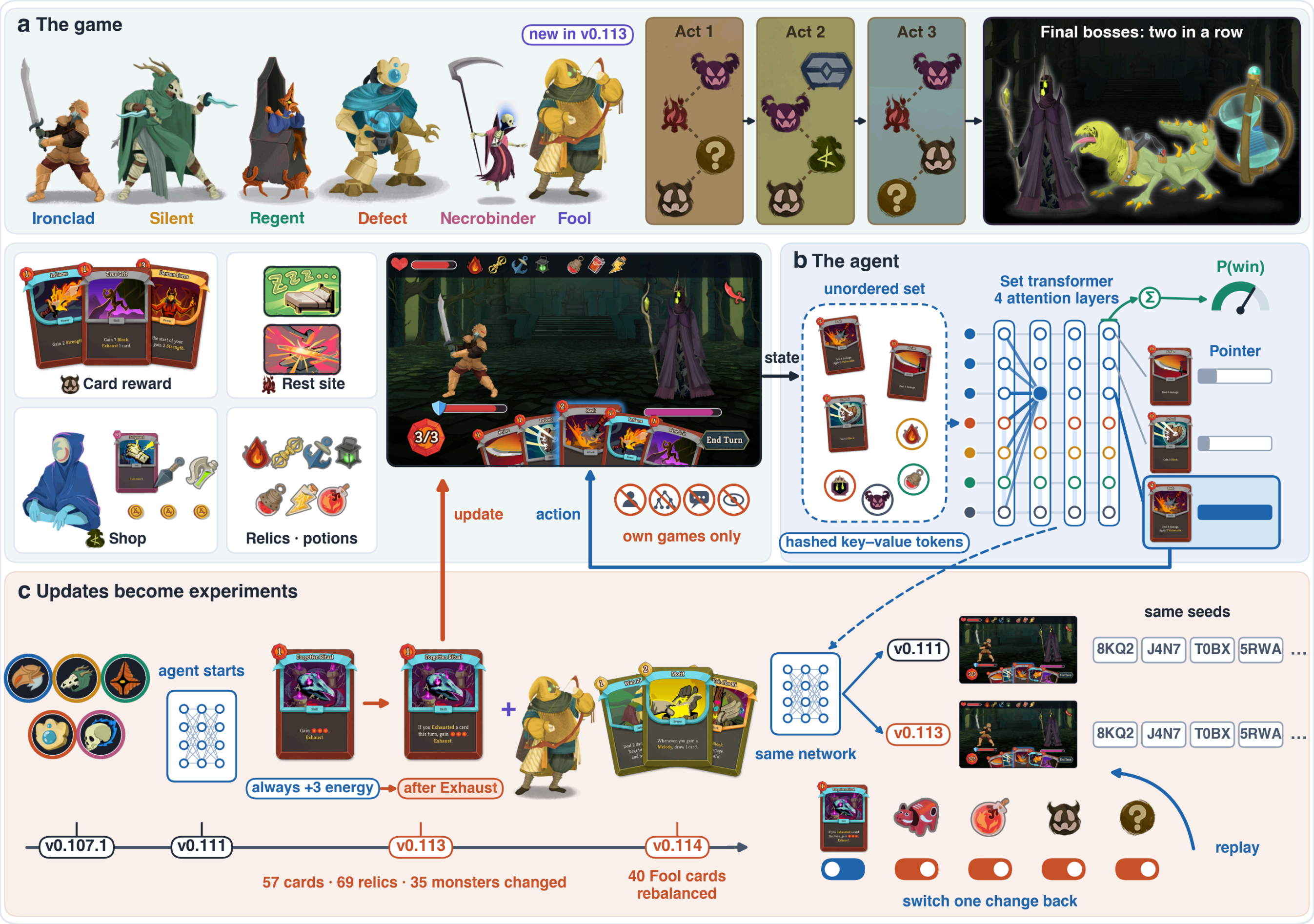

**Figure 1 | A changing game, a small agent and updates as experiments. a**, *Slay the Spire 2*: six characters, three acts of fights, shops, rest sites and events, and two final bosses in a row. **b**, SpireZero reads the visible game state as an unordered set of tokens, scores every legal action and estimates its chance of winning; it learns only from its own games. **c**, Each update changes a known list of rules. Playing the same games before and after an update measures its effect, and switching single changes back shows where that effect comes from.

At the highest difficulty (Ascension 10), SpireZero wins between 70% and 75% of fresh games with each of the five original characters, 73% on average (Fig. 2a). The same networks, moved without retraining to the older version 0.107.1 on which language models have been benchmarked, still win 60% of games; the strongest language-model agent wins 35% there [9] (Fig. 2b). When an update added a sixth character, the Fool, a further copy of the agent learned to win more than half of its games with it (Fig. 2a).

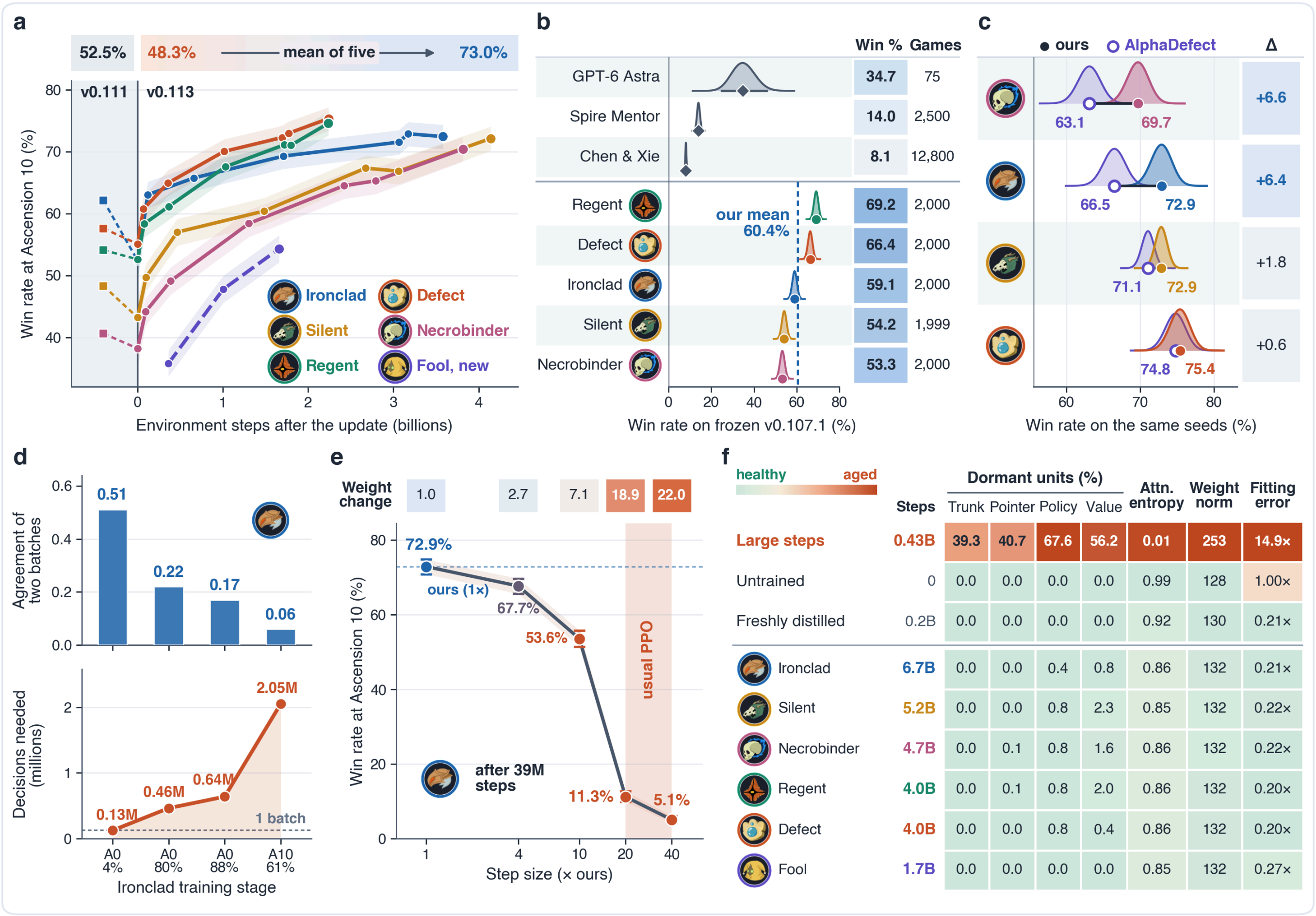

**Figure 2 | SpireZero matches an engineered agent and keeps learning because its steps are small. a**, Win rate at the highest difficulty for each character across the update to version 0.113. **b**, The same networks on the older version used to benchmark language models, without retraining, next to published agents [9, 12, 13]. **c**, SpireZero and AlphaDefect on the same games, both without search. **d**, Agreement between the learning directions of two independent batches at four stages of training (top), and the amount of experience needed for a reliable direction (bottom). **e**, Win rate of one agent after 40 million decisions of further training at increasing step sizes. **f**, Signs of network ageing in an early agent trained with large steps, in untrained and freshly distilled networks, and in the six agents trained with small steps. Shaded bands and bars, 95% intervals; 2,000 games per SpireZero point in **a**, **b** and **e**. In **e**, the learning rate reached its ceiling for the two largest step sizes (Methods).

A concurrent study provides an independent reference point. AlphaDefect [11] is also trained with PPO on the game's own code, but shows its network the engine's preview of each action's outcome, trains its value network on information hidden from the player and adds tree search at decision time. Playing exactly the games on which AlphaDefect was evaluated, without search on either side, SpireZero is ahead by about six percentage points with two characters and level with the other two for which game-by-game results were published (Fig. 2c). A plain network learning from its own experience is therefore enough to reach this level.

## Small steps keep the agent learning

SpireZero keeps improving over billions of decisions, where standard recipes stall, because it allows for how noisy each learning step is. We compared the directions in which two independent batches of experience would move the policy. The two directions agree less and less as the agent improves, and for a skilled agent at the highest difficulty they are almost unrelated, so that a single batch, although it holds more than 100,000 decisions, carries little signal [14, 15] (Fig. 2d). Over consecutive learning steps the weights of the policy accordingly follow a path that is nearly indistinguishable from a random walk (Extended Data Fig. 1a). Improvement therefore accumulates only over hundreds of learning steps.

Large steps destroy this slow accumulation. We took an Ironclad agent that won 73% of its games and continued its training with progressively larger steps. Over 40 million further decisions our usual step size preserved the win rate, larger steps eroded it progressively, and step sizes typical for PPO reduced it to between 5% and 11% (Fig. 2e).

Small steps also prevent the network from ageing. An early agent that we trained from scratch with larger steps showed every reported sign of plasticity loss [3–5]: nearly half of its units went dormant, its attention collapsed onto a few tokens, its weight norm doubled and it became far worse than an untrained network at fitting new targets [16] (Fig. 2f). After copying that agent's behaviour once into a fresh network by distillation [17, 18], we trained all six characters with small steps only. Billions of decisions later, none of the six networks shows any of these signs: no unit in the Transformer layers is dormant, attention remains broad, the weight norm has not grown and each network fits new targets as readily as a freshly distilled one (Fig. 2f; Extended Data Fig. 1b, c). The same holds in practice. After a real update, a network that had already trained on 3.5 billion decisions adapted no more slowly than a freshly distilled copy of itself (Extended Data Fig. 1d). Long training does not by itself age a network; large, noisy steps do.

## Two real updates, one absorbed in hours and one not

Two updates arrived while SpireZero was training, and they differed sharply in structure (Fig. 3).

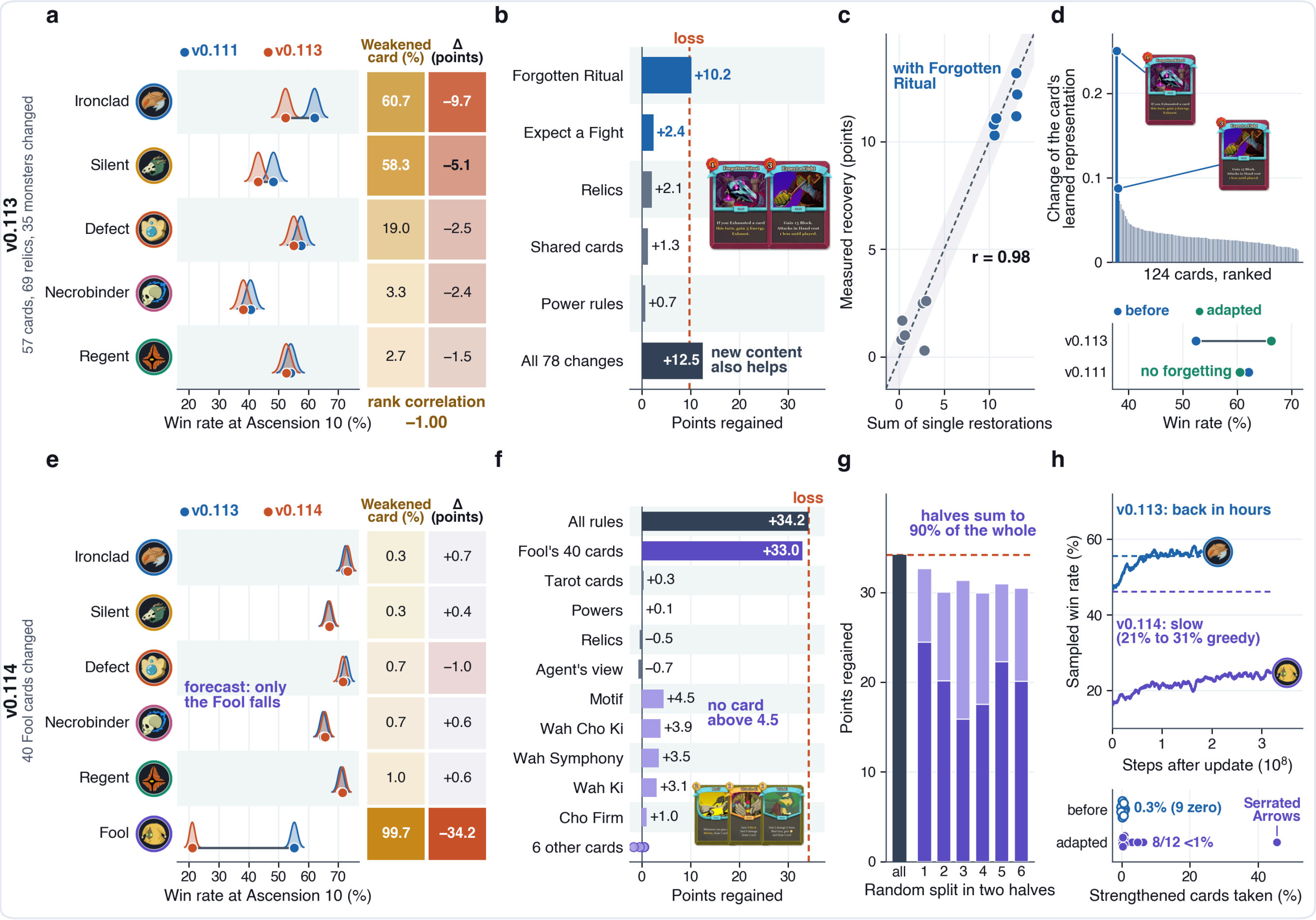

**Figure 3 | Both updates can be taken apart rule by rule, but only one is repaired in hours. a–d**, Version 0.113. **a**, Win rate of each agent before and directly after the update, with the share of its earlier decks that contained a weakened card. **b**, Points regained by the Ironclad when changes are reverted. **c**, Twelve artificial updates built from random halves of the real one: measured recovery against the sum of single-change effects. **d**, Change in each card's learned representation after adaptation (top), and win rate before and after adaptation on the new and the old version (bottom). **e–h**, Version 0.114. **e**, Win rates before and after the update; the forecast was recorded before any game was played. **f**, Points regained by the Fool when changes are reverted. **g**, Six random splits of the update into two halves. **h**, Recovery after each update, with actions sampled as in training (top), and how often the Fool takes the 12 strengthened cards before the update and after adaptation (bottom).

The first, version 0.113, arrived midway through training, when the agents won roughly 40% to 60% of their games. It changed 57 cards, 69 relics and 35 monsters. Played without any retraining, the five agents lost up to 10 percentage points, and the agents that had relied most on weakened cards lost most (Fig. 3a). We then reverted the changes one at a time. For the Ironclad, almost the entire loss came from a single card. *Forgotten Ritual* used to grant energy (the budget for playing cards) whenever it was played; after the update it does so only if another card has been exhausted (removed from the fight) in the same turn. Restoring this one card returned all 10 lost points, and the remaining 77 changes together mattered little (Fig. 3b). When we assembled 12 artificial updates from random halves of the real one, the damage of each was predicted by adding up the effects of the single changes it contained ($r = 0.98$; Fig. 3c).

The agent's repair was quick and just as local. Continuing training with the usual small steps recovered the lost level within three to five hours, about 3% of the experience that training from scratch had required, whereas larger learning steps slowed the recovery (Extended Data Fig. 1d). Inside the network, the two representations that changed most were those of *Forgotten Ritual* and *Expect a Fight*, the two Ironclad cards that the update had altered (Fig. 3d, top). The repair left the rest of the agent's play intact: returned to the old version of the game, it played as well as the agent that had never left it (Fig. 3d, bottom). What it gave up was the changed card itself, a point to which we return below.

The second update, version 0.114, arrived when this analysis was complete, which allowed a forecast. Before running a single game on the new version, we read the list of changes, saw that nearly all of them concerned the Fool, and recorded our prediction: the five original characters would be unaffected and the Fool would fall. The five indeed moved by less than one percentage point on average, whereas the Fool dropped from 55% to 21% (Fig. 3e). Reverting the changes accounted for the whole loss, as it had for the Ironclad, and random halves of the update again added up to nearly the whole (Fig. 3f, g). This time, however, the damage was spread across 40 rebalanced cards, and no single card accounted for more than a few points (Fig. 3f).

The agent did not shake off this widely spread update in hours. After 300 million further decisions the Fool had regained 10 of the 34 lost points, a pace no faster than that at which it first learned the character (Figs. 2a and 3h). It stopped taking several cards that the update had weakened and adopted one that it had strengthened, but eight of the 12 strengthened cards, almost never taken before, are still almost never taken (Fig. 3h). In both cases the loss could be attributed rule by rule, but only the concentrated one was repaired by a few hours of training.

### An economy learned from winning

Before turning to what SpireZero overlooks, we ask what the outcome of a game can teach across a run of some 50 floors. The agents do not play short-sightedly: from a reward that says only how far a run got and whether it was won, they developed a consistent way of handling every resource in the game (Fig. 4). The bonuses for climbing floors and defeating bosses are dispensable: an agent trained without them for nearly a billion decisions plays at the same level (Extended Data Fig. 2c).

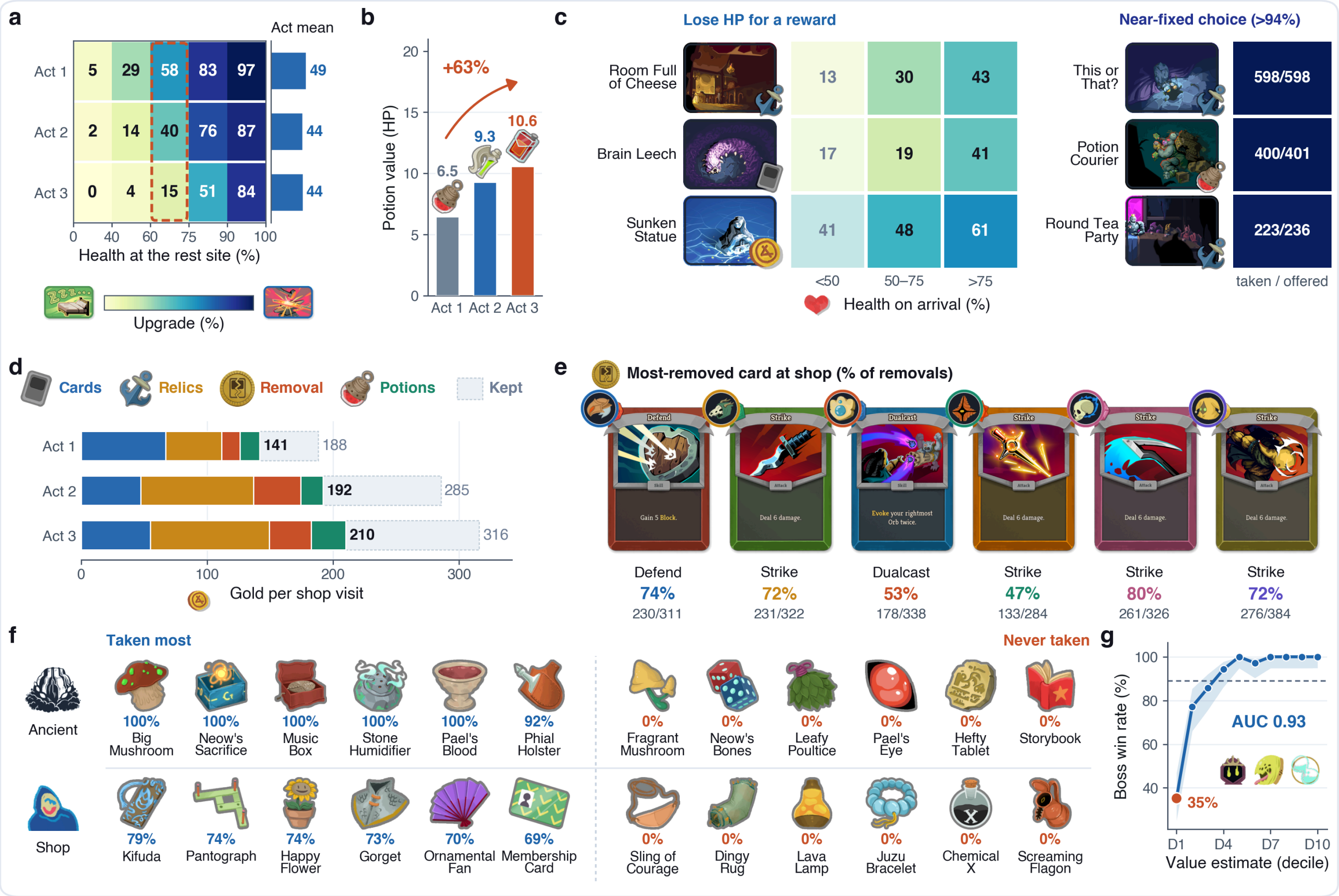

**Figure 4 | An economy learned from winning.** **a**, Share of rest-site visits used to upgrade a card, by act and health. **b**, Value of one potion in health points, by act. **c**, Events that trade health for a reward, by health on arrival (left), and events with a near-fixed choice (right). **d**, Gold spent per shop visit, by act. **e**, The starting card that each character most often pays to remove. **f**, Relics taken most often and relics never taken, from Ancients (figures who offer a relic at the start of an act) and in shops. **g**, Win rate against the final bosses by the agent's own estimate on entering.

At rest sites, where a player must choose between healing and upgrading a card, the agents heal when they are hurt and upgrade when they are healthy, and they become more cautious before the final bosses (Fig. 4a). This contrasts with language-model agents playing the original game, which heal far more often than strong players do [19]. The value that the agents attach to a potion, measured in the health they would give up for it, rises by more than half from the first act to the third, as the remaining fights become fewer and harder (Fig. 4b). The same pricing appears in events that offer a reward in exchange for health: the healthier the agent, the more often it accepts, while offers that are simply good are taken almost every time (Fig. 4c).

Spending follows a plan as well. In shops the agents buy cards early and relics later (Fig. 4d). Without being told that a deck with fewer weak cards is stronger, each character learned which of its starting cards to pay to remove (Fig. 4e). Some relics are taken at every opportunity and others never (Fig. 4f).

Underlying these choices is a well-calibrated value estimate: on entering the final bosses, it separates the runs that will be won from those that will be lost with an area under the curve of 0.93 (Fig. 4g).

## Where SpireZero and strong players part ways

An agent that prices health, gold and its own deck dozens of floors ahead does not lack a long horizon. How, then, does its self-taught style compare with that of strong human players? Across some 400 cards, SpireZero's preferences are positively but only moderately correlated with those of strong players [20] (rank correlations between 0.18 and 0.58 across characters; Fig. 5a, c). The disagreements are systematic. Cards that are useful on their own SpireZero takes as often as strong players do, and frequently more often. Cards that pay off only in combination with others, the centrepieces of what players call archetypes, it almost never takes, although strong players take the best known of them in a third to a half of their opportunities (Fig. 5a, b). SpireZero's own losses fit this picture. Two thirds of them come in the two final boss fights: losing runs enter in good health, but their weaker decks cost them far more health in the first fight and leave too little for the second (Extended Data Fig. 2a, b).

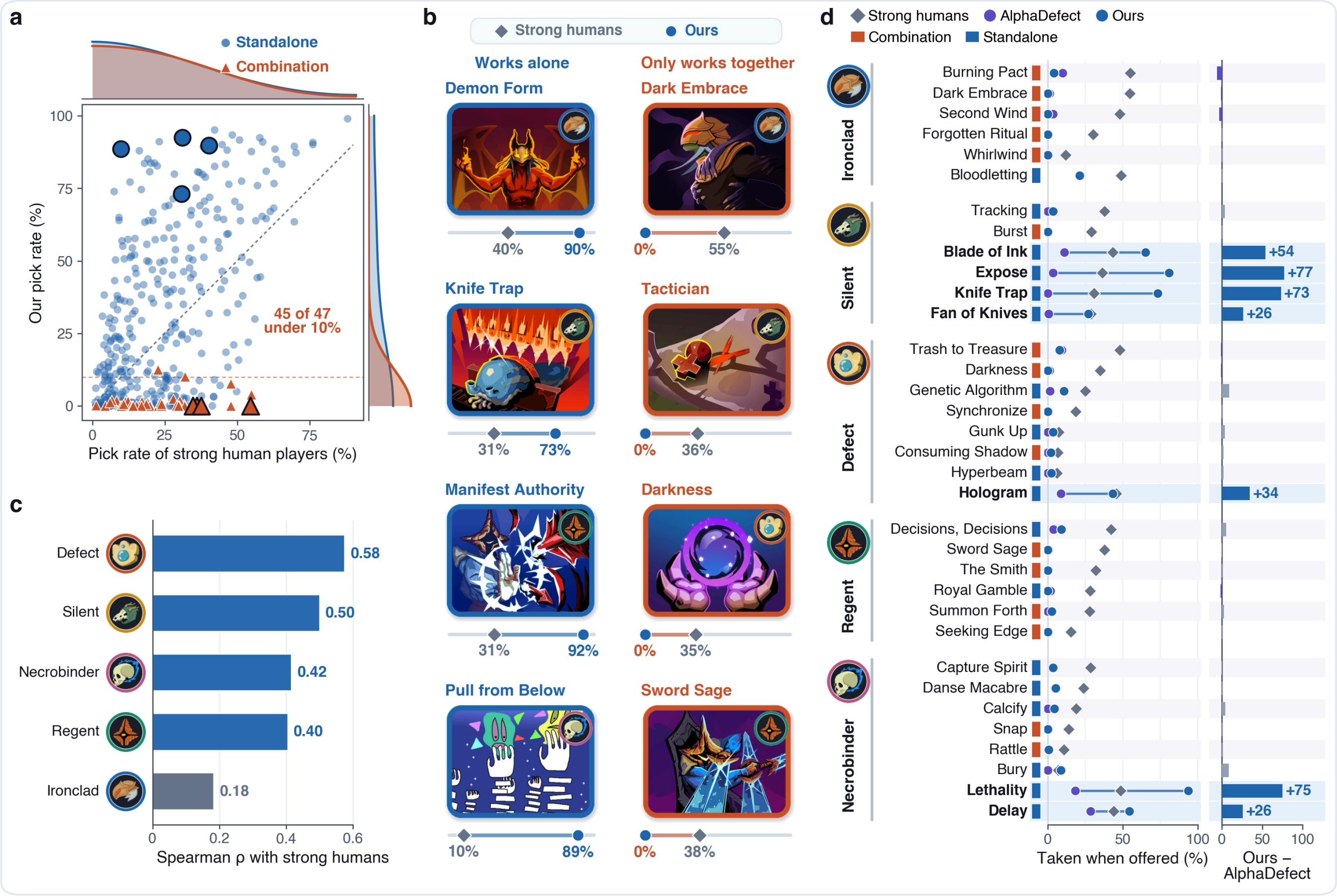

**Figure 5 | Two independent agents overlook the same cards: those that need partners.** **a**, How often SpireZero and strong players take each of about 400 cards; triangles mark cards that pay off only in combination. **b**, Four cards that work alone and four that need partners. **c**, Rank correlation between SpireZero's and strong players' preferences for each character. **d**, The 34 cards that AlphaDefect takes least in comparison with humans, with the preference of each party and the difference between the two agents.

This gap does not stem from our particular design. AlphaDefect published the cards that its agents take least in comparison with humans [11]. SpireZero neglects most of the same cards, and every card on the list that needs partners is neglected by both agents. Where the two agents differ, the card is one that works alone, and each has developed tastes of its own (Fig. 5d). The two systems differ in network, inputs, value training and search, so their agreement on the cards that need partners is unlikely to be a property of either design.

## Quick to abandon, slow to rediscover

The first update happened to push one card across the boundary between cards that work alone and cards that need partners, and so provided a natural experiment (Fig. 6). Before the update, *Forgotten Ritual* worked alone and the Ironclad agent took it in three of every four opportunities. After the update it needed a partner. The agent did not learn the new pairing. It stopped taking the card, and its interest halved within about 35 million decisions and soon reached zero (Fig. 6a, b).

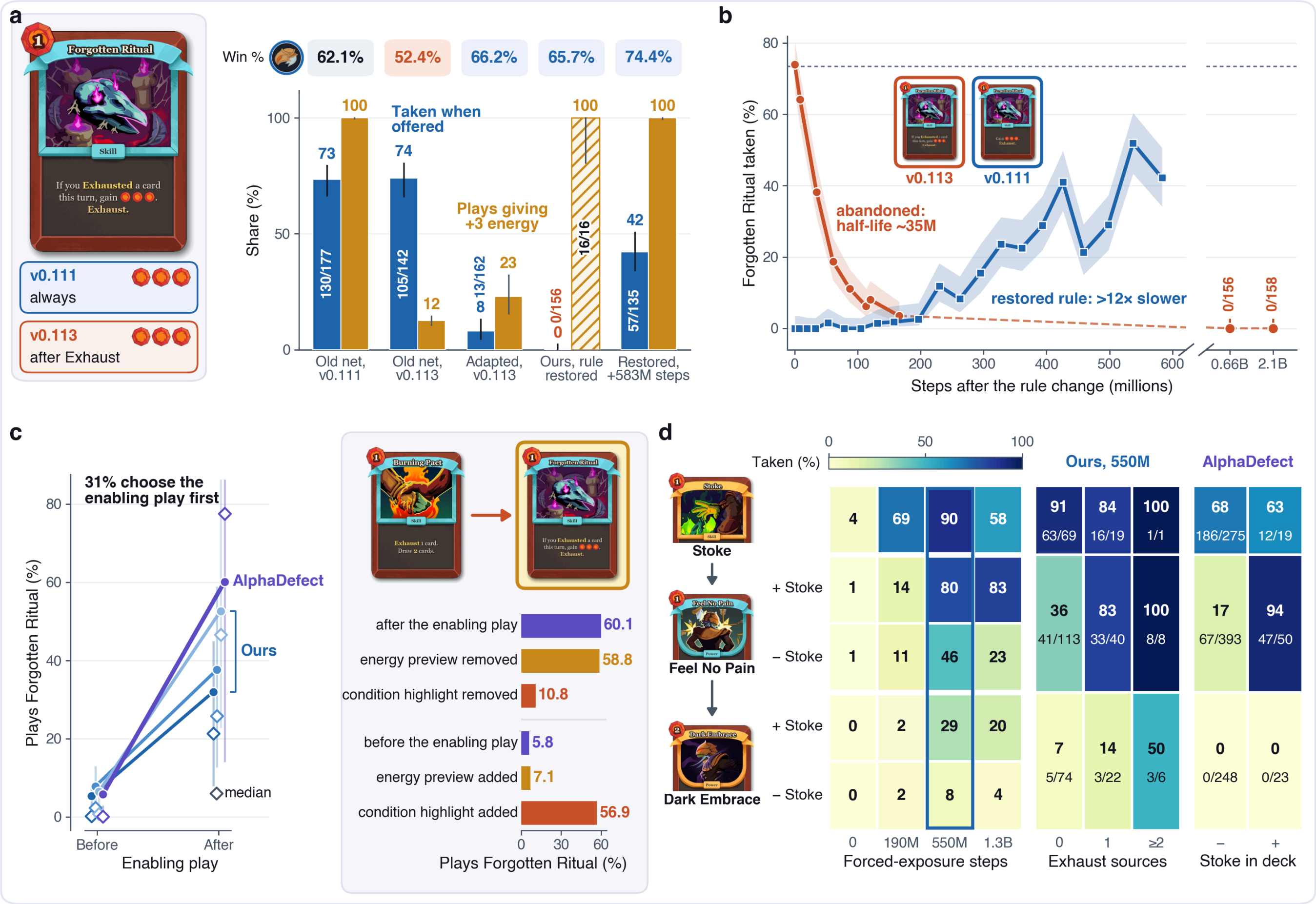

**Figure 6 | Quick to abandon, slow to rediscover. a**, *Forgotten Ritual* in five situations: how often the Ironclad agent takes it, how often playing it yields energy, and the agent's win rate. **b**, How often the agent takes the card after it was weakened and after its original rule was restored. **c**, How often three SpireZero networks (the main agent, a control and the agent of **d**) and AlphaDefect play the card before and after the enabling play is made for them (left), and the effect of altering AlphaDefect's inputs (right). **d**, How often an agent made to take exhaust cards in half of its training games chooses them in its other games, over training and by what its deck already holds; the released AlphaDefect network, split by whether its deck holds *Stoke*, for comparison.

We then gave the card back its original rule. The adapted agent, offered the restored card 156 times, did not take it once. Only training brought it back, and slowly. For roughly 200 million decisions the card stayed untouched, and its return began with the few occasions on which the game itself placed a copy in the agent's hand during combat. From this trickle of experience the preference recovered, reaching about half of its former level after 400 to 500 million decisions (Fig. 6b). Rediscovery was more than 10 times slower than abandonment. This asymmetry is expected of an agent that learns only from the options it tries, and resembles the hot-stove effect described for human learning [21].

The agent's failure with the changed card does not reflect a misunderstanding of its rule. In combat positions where the agent held *Forgotten Ritual* together with a card that could enable it, it rarely wasted the ritual by playing it too early, and once we made the enabling play on its behalf, it often followed with the ritual. What it seldom did was make the enabling play itself. AlphaDefect behaved in the same way, and manipulating its inputs showed that its decision followed the game's own on-screen signal that the condition was met rather than its engineered preview of the outcome (Fig. 6c). Both agents recognize a satisfied condition; neither takes the earlier step that would create it.

The barrier can nevertheless be crossed when experience of a whole chain of related cards is supplied. In half of an agent's training games we made it take every card of the exhaust family whenever one was offered, and kept these forced choices out of the policy's learning signal. In its other games the agent then began to choose these cards for itself, and in the order of their dependence: first *Stoke*, which works alone, then *Feel No Pain*, which needs it, then *Dark Embrace*, which needs both (Fig. 6d). The dependence between cards can be learned once its first link is in hand. Learning from one's own experience stalls at the first link, where taking a card brings no benefit until its partners arrive.

## Discussion

Software that evolves through documented patches turns continual reinforcement learning into an experimental science [2, 7]. Changes do not need to be invented, because the developers supply them, and every change can be undone. The two updates studied here support the view that a real change alters only a few underlying mechanisms [22, 23], and they also qualify it. In both, the damage to the character hit hardest could be attributed to individual rules and added up across them, and knowing what a policy relies on was enough to say in advance who would be hurt [8, 24]. But sparse attribution did not guarantee quick repair. When one card carried the loss, the agent routed around it within hours; when 40 cards moved together, it had to relearn the character at the pace of first learning.

Our results also connect three problems that are usually treated separately: the plateaus of policy-gradient methods [14, 15], the loss of plasticity under long training [3–5] and forgetting after a change [6]. In SpireZero all three yield to the same remedy. The more skilled the agent, the noisier each learning step becomes, and steps of customary size then both waste the signal and damage the network, which may help to explain why OpenAI Five's long-trained agent was overtaken by a rerun from scratch [25, 26]. With small enough steps, the same network kept learning over seven billion decisions, adapted to a new version as fast as a fresh one, and retained the old version, in line with recent accounts of why on-policy learning forgets little [6, 27].

Finally, comparing game versions, and our agent with an independent one, shows the reach and the limit of learning from experience. From the outcome of its games alone, SpireZero learned to price health, gold, potions, cards and relics, and it plays at the level of a far more heavily engineered agent. It stops where an investment pays nothing until a later, separate choice completes it: retrying an option it has learned to avoid, or collecting cards that are worthless until their partners arrive [21, 28–30]. Because supplied experience unlocks such combinations in order, we expect such combinations to be reached by exploration that commits to a plan across many decisions.

These findings are not specific to card games. Any agent that keeps learning inside software that is revised around it faces the same questions. Our results suggest that small learning steps are a sound default for an agent meant to learn for years; that a local change to its world can be absorbed on the fly, whereas a broad rebalancing should be budgeted like learning anew; and that tools or options which once failed and were later fixed will stay unused unless the agent is made to try them again.

## Methods

**Environment and fidelity.** The agent plays the game's own compiled library, hosted in a process without graphics. Only the presentation layer is replaced, and a small number of code changes skip animations and keep execution deterministic. Every decision that the released client offers a player is offered to the agent as a list of legal options. To verify fidelity, 129 runs recorded in this environment (11,502 decisions, covering all six characters, the lowest and highest difficulties and rarely seen screens) were replayed step by step inside the released client through a verification add-on; observations and legal options agreed at every step. The same environment was built for four versions of the game (0.107.1, 0.111, 0.113.1, called 0.113 in the text, and 0.114), each of which passed the same suite of determinism and regression tests. Unless stated otherwise, results are for the highest difficulty, Ascension 10, played without reloading saved games.

**Observation and network.** The observation contains what a player with perfect memory could know: everything currently on screen, the composition but not the order of the card piles, and the history of the current fight. Hidden information, such as the order of the draw pile, the contents of unvisited rooms and unopened rewards, is excluded. Each entity (card, relic, potion, enemy, map node and so on) is encoded as a token from its name and attributes. Names are mapped by hashing into a shared embedding table, so that new content introduced by an update needs no change to the network. A four-layer Set Transformer without positional encoding processes the tokens, a pointer layer scores the legal options, and a value head estimates the expected reward of the run. Sizes and hyperparameters are listed in Extended Data Table 1.

**Training.** Agents are trained with PPO [10] and generalized advantage estimation [31], with a second, slower timescale for deck-building decisions, in which a whole fight between two such decisions counts as a single step, so that credit reaches choices whose consequences appear many fights later. With no discounting, the temporal-difference error of decision $t$ is $\delta_t = r_t + V(s_{t+1}) - V(s_t)$ and the advantage is

$$A_t = \delta_t + \lambda_t A_{t+1}.$$

For decisions in combat, $\lambda_t = \lambda$ at every step. For deck-building decisions, $\lambda_t = \lambda_\text{b}$ when decision $t + 1$ is again a deck-building decision and $\lambda_t = 1$ otherwise, so that the advantage decays once per deck-building decision and not once per card played. The reward is 1 for winning a run, with small bonuses for each floor climbed and each act boss defeated; the win-only ablation removes the bonuses. Each learning step uses a large batch of fresh experience once and maximizes the clipped objective of PPO,

$$L = \mathbb{E}_t\Big[\min\big(\rho_t A_t,\ \mathrm{clip}(\rho_t, 1-\epsilon, 1+\epsilon)A_t\big) - c_v\big(V(s_t) - G_t\big)^2 + c_e H\big(\pi(\cdot \mid s_t)\big)\Big],$$

where $\rho_t$ is the ratio of the new to the old probability of the chosen action, $G_t$ the return target and $H$ the entropy of the policy (constants in Extended Data Table 1). The learning rate is adjusted continuously to hold the change of the policy per learning step, measured as a Kullback–Leibler divergence, at a target of 0.0005, about 20 times smaller than usual. After every learning step the learning rate is multiplied by

$$\min\left(1.03,\ \max\left(\frac{1}{1.03},\ \sqrt{D^*/\bar{D}}\right)\right),$$

where $D^*$ is the target and $\bar{D}$ a running average of the measured divergence, and it is kept between $10^{-6}$ and $10^{-3}$. Training began from scratch at the lowest difficulty; the resulting agent was distilled once into a freshly initialized network [17, 18], and all later training took place at the highest difficulty. The Ironclad was trained alone and the other characters branched from a network trained on all five original characters. No human data were used at any stage. Each agent was trained on one graphics processor (two for the Ironclad) with 160 to 256 copies of the game running in parallel, at 3,000 to 8,000 decisions per second, so that a billion decisions took between one and a half and four days.

**Evaluation.** Unless stated otherwise, a win rate is measured on 2,000 games with random seeds that were never used in training, with the agent choosing its most probable action. Evaluated weights are the average of four consecutive saved checkpoints. The rare game that the environment aborted was excluded (at most one per evaluation). Because a seed determines a game, two agents, or one agent under two rule sets, can be compared on the same games; such paired comparisons were always run on the same machine and tested with McNemar's test, $z = (b - c)/\sqrt{b + c}$, where $b$ and $c$ count the games won by only one of the two. Intervals in the figures are 95% Wilson intervals. For the comparison with AlphaDefect [11], SpireZero played the seed lists published by its authors and was paired, seed by seed, with their published results for play without search. This was possible for four characters (1,000 games each for the Ironclad, the Defect and the Necrobinder, and 3,000 for the Silent), all on version 0.113.1.

**Noise of learning steps.** For two independent batches of $N$ decisions collected with the same network weights, with policy gradients $g_1$ and $g_2$, the agreement and the batch size at which signal equals noise are

$$c = \frac{g_1 \cdot g_2}{\|g_1\|\,\|g_2\|}, \qquad N_1 = N\frac{1-c}{c}.$$

The straightness of the weight path over a window of $k$ learning steps is

$$S_k = \frac{\|\theta_{t+k} - \theta_t\|}{\sum_{i=t}^{t+k-1} \|\theta_{i+1} - \theta_i\|},$$

which equals 1 for a straight path and $1/\sqrt{k}$ for a random walk.

**Step-size experiment.** Five copies of one Ironclad agent were trained for 300 further learning steps with different targets for the policy change per step, from our usual value to 40 times that value, and evaluated on the same 2,000 games. For the two largest targets the learning rate reached its permitted maximum, so that the realized policy change per step was about 16 times our usual value.

**Plasticity measurements.** For each network we recorded, on 4,096 game states of its own character, the fraction of dormant units [5], the entropy of attention in the trunk, the norm of the weights (without the lookup table) and the error remaining after 1,000 steps of fitting random targets [16]. The last quantity is reported relative to that of an untrained network fitted on the same states. A unit is counted as dormant when its mean absolute activation $a_i$, relative to the mean over the $n$ units of its layer, satisfies $a_i/\left(\frac{1}{n}\sum_j a_j\right) \le 0.1$. The early agent trained with large steps had the same network and conventional PPO settings, with a policy change per learning step about 30 to 40 times larger.

**Reverting and forecasting updates.** For each update, the list of changes was obtained by comparing the decompiled code of consecutive versions. We implemented switches that restore the earlier behaviour of each changed card, relic, potion or rule (78 switches for version 0.113, covering every changed card, relic, potion and rule that concerns the five original characters, with most changes to monsters and all new content left in place; 66 switches for version 0.114, covering its changes to the standard game) and measured the effect of each on the same games. Artificial updates were built from random subsets of the switches. The forecast for version 0.114, with its criteria, was committed to our repository before any agent had played that version.

**Analyses of behaviour.** Preferences were computed from 300 recorded games per character. The price of a potion was obtained by regressing the change in the agent's value estimate across a fight on the health and potions spent in it: with $\Delta V = \beta_{\mathrm{h}}\,\Delta\text{health} + \beta_{\mathrm{p}}\,\Delta\text{potions}$, the price of one potion in health points is $\beta_{\mathrm{p}}/\beta_{\mathrm{h}}$. Cards were labelled as paying off only in combination from their rules text and from community guides to the game's archetypes; the labels are listed with the figure data. Human preferences are those of players with a win rate above 75% in a public database of community runs [20]. Changes in the representation of a card were measured as the distance between its embeddings before and after adaptation. In the combat probe we collected positions in which the agent held *Forgotten Ritual* and a card able to enable it, and compared its choices before and after the enabling card was played for it; the same positions were presented to the released AlphaDefect network, whose input channels were additionally altered one at a time. In the supplied-experience experiment, cards of the exhaust family were added to the deck whenever they were offered in half of the training games, and these forced choices contributed no gradient to the policy.

**Data and code availability.** The code and the figure data are available at https://github.com/void-echo/sls2-rl; the remaining source data and the trained networks will be added on publication. The game is commercial software and is not redistributed.

## Extended Data

**Extended Data Table 1 | Network and training.**

| Component | Setting |
|---|---|
| Identity embeddings | Hashed table, 65,536 × 256 (16.8 million weights) |
| Network | Four-layer Set Transformer, width 256, eight heads; pointer layer; value head (5.1 million weights) |
| PPO | 131,072 decisions per learning step, one pass in minibatches of 16,384, Adam, clipping $\varepsilon = 0.2$, value weight 0.5, entropy bonus 0.01, no discounting, $\lambda = \lambda_b = 0.95$ |
| Step size | Policy change per learning step held at a KL divergence of 0.0005 by adjusting the learning rate (at most 0.001) |
| Hardware | One graphics processor per agent (two for the Ironclad), 160–256 parallel games, 3,000–8,000 decisions per second |
| Reward | 1 for a win, 0.01 per floor, 0.2 per act boss |
| Experience | 1.7 to 7.1 billion decisions per character, more than 29 billion in total |

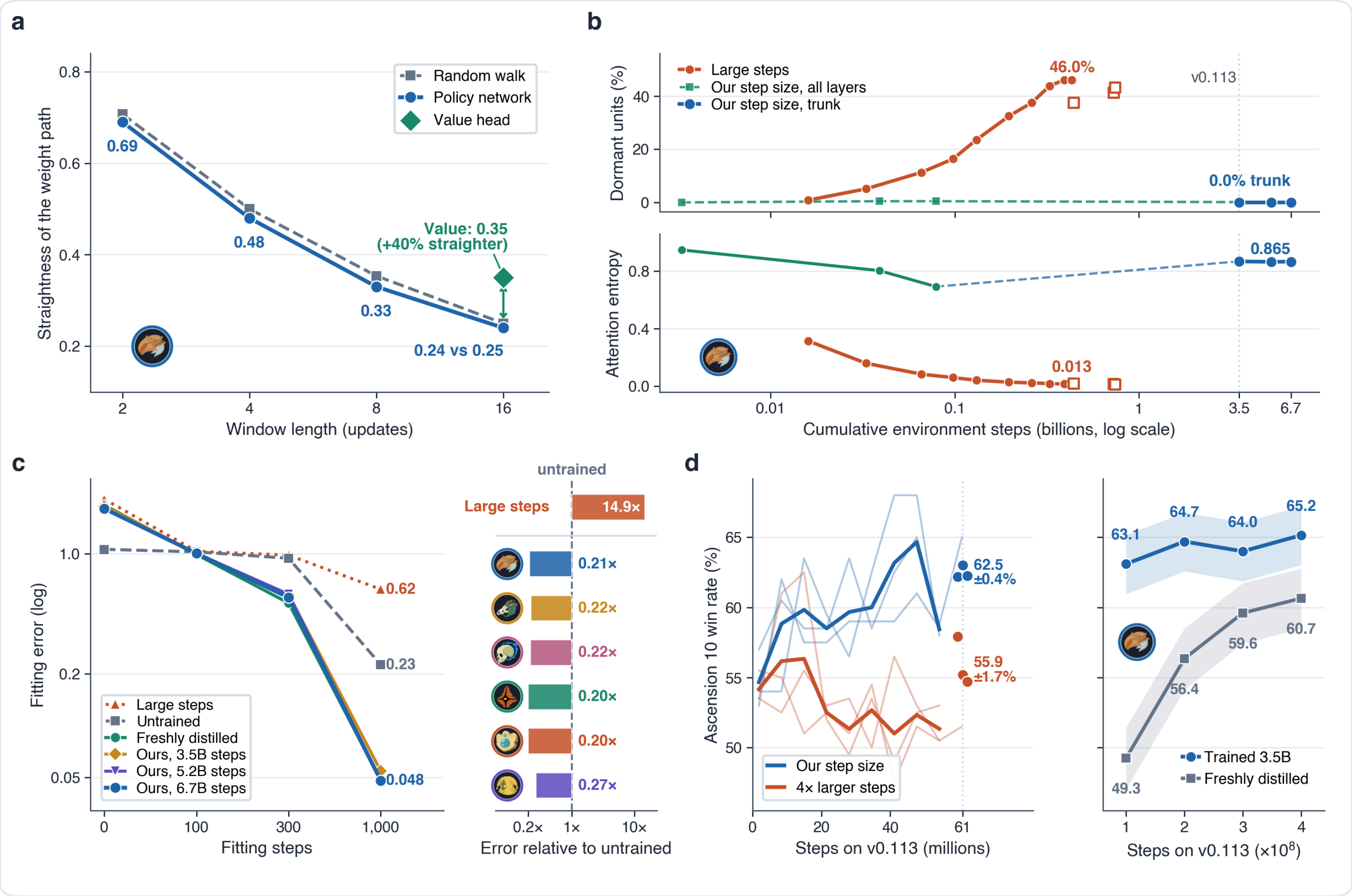

**Extended Data Fig. 1 | Noise, ageing and step size. a**, Straightness of the path of the policy weights over consecutive learning steps, compared with a random walk. **b**, Dormant units and attention entropy during training with large and with small steps. **c**, Fitting random targets: error during fitting (left; the agent trained with large steps was measured on states of an earlier version, on which an untrained network reaches 0.042), and final error relative to an untrained network on the same states (right). **d**, Recovery after the update to version 0.113 with small and with larger steps (left), and for a long-trained network against a fresh copy (right).

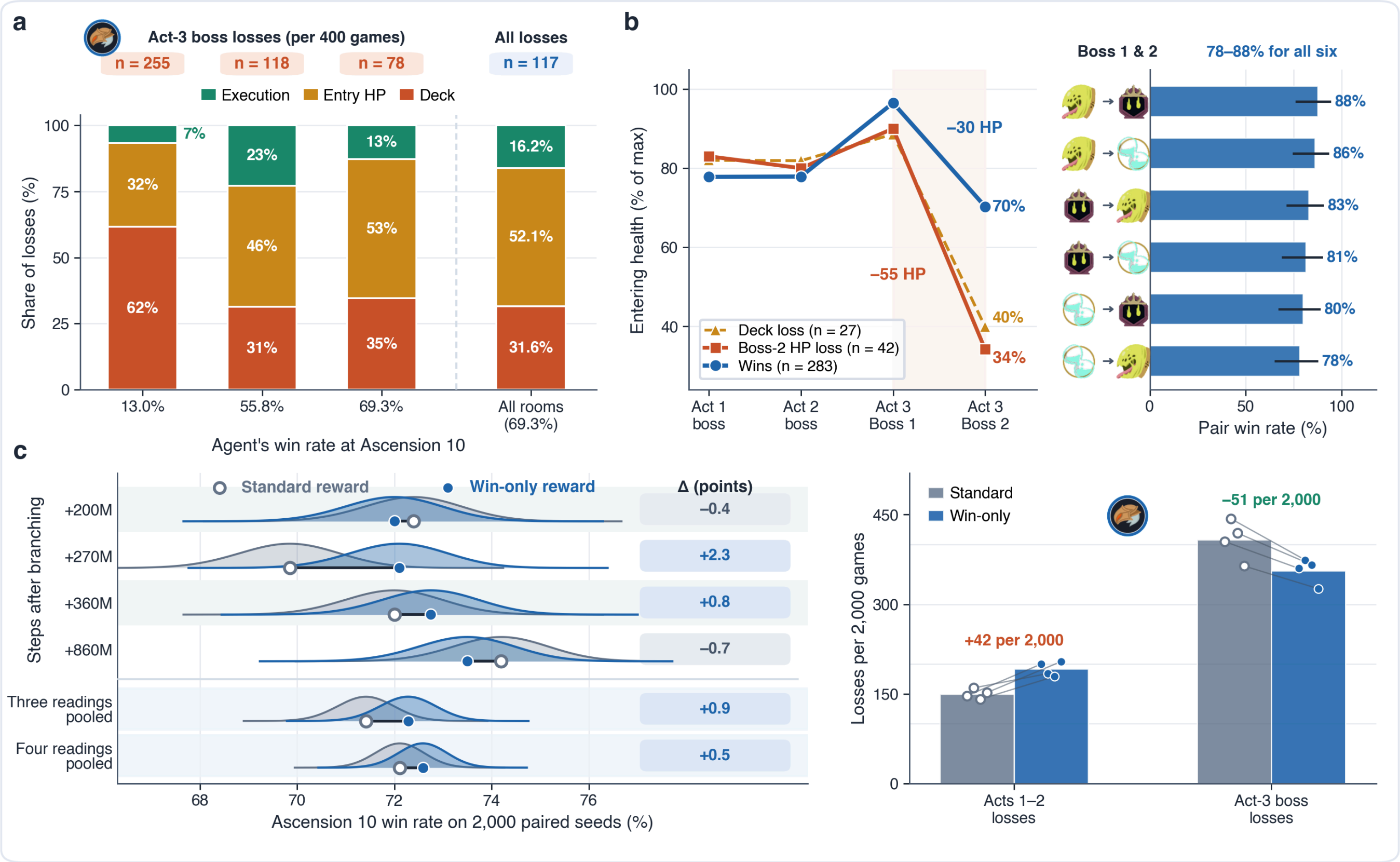

**Extended Data Fig. 2 | Where runs are lost, and training on wins alone. a**, Losses to the final bosses in three generations of the Ironclad agent, classified by replaying each lost fight from its start with the same and with full health. **b**, Health on entering each boss fight for won and lost runs, and win rate for each pair of final bosses. **c**, An agent trained on wins alone against its control on the same games at four times after the fork, and the stage at which each loses.